\documentclass[conference]{IEEEtran}
\makeatletter\if@twocolumn\PassOptionsToPackage{switch}{lineno}\else\fi\makeatother
\usepackage{bigdelim}

\usepackage{graphicx}
\usepackage[T1]{fontenc}
\usepackage[numbers,sort&compress]{natbib}
\usepackage{mathtools}
\usepackage{booktabs}
\usepackage{lineno,hyperref}
\usepackage{amsmath}
\usepackage{amsfonts}
\usepackage{amssymb}
\usepackage{amsmath}
\usepackage{multirow,morefloats,floatflt,cancel,tfrupee}
\usepackage{makecell}
\usepackage{tikz}
\usepackage{setspace}
\usepackage{color}
\usepackage{multirow}
\usepackage{adjustbox}
\usepackage{bbding}
\usepackage{physics}
\usepackage{colortbl}
\usepackage{xcolor}
\usepackage{pifont}
\usepackage[nointegrals]{wasysym}
\renewcommand\IEEEkeywordsname{Keywords}

\makeatletter

\let\old@ps@IEEEtitlepagestyle\ps@IEEEtitlepagestyle
\def\confheader#1{%
    \def\ps@IEEEtitlepagestyle{%
        \old@ps@IEEEtitlepagestyle%
        \def\@oddhead{\strut\hfill#1\hfill\strut}%
        \def\@evenhead{\strut\hfill#1\hfill\strut}%
    }%
    \ps@headings%
}
\makeatother

\confheader{%
    \parbox{17cm}{\centering 13th International Conference on Computer and Knowledge Engineering (ICCKE 2023), November 1-2, 2023, Ferdowsi University of Mashhad}
}

\begin{document}

        \title{Traffic Sign Recognition Using Local Vision Transformer}
%


\author{\IEEEauthorblockN{Ali Farzipour,
Omid Nejati Manzari,
Shahriar B. Shokouhi}\\
School of Electrical Engineering, Iran University of Science and Technology, Tehran, Iran}

\maketitle

\begin{abstract}

Recognition of traffic signs is a crucial aspect of self-driving cars and driver assistance systems, and machine vision tasks such as traffic sign recognition have gained significant attention. CNNs have been frequently used in machine vision, but introducing vision transformers has provided an alternative approach to global feature learning. This paper proposes a new novel model that blends the advantages of both convolutional and transformer-based networks for traffic sign recognition. The proposed model includes convolutional blocks for capturing local correlations and transformer-based blocks for learning global dependencies. Additionally, a locality module is incorporated to enhance local perception. The performance of the suggested model is evaluated on the Persian Traffic Sign Dataset and German Traffic Sign Recognition Benchmark and compared with SOTA convolutional and transformer-based models. The experimental evaluations demonstrate that the hybrid network with the locality module outperforms pure transformer-based models and some of the best convolutional networks in accuracy. Specifically, our proposed final model reached 99.66\% accuracy in the German traffic sign recognition benchmark and 99.8\% in the Persian traffic sign dataset, higher than the best convolutional models. Moreover, it outperforms existing CNNs and ViTs while maintaining fast inference speed. Consequently, the proposed model proves to be significantly faster and more suitable for real-world applications.
\end{abstract}


\section{Introduction}

The importance of safety systems in self-driving cars and driver assistance systems is obvious. These safety systems are an essential part of vehicles in terms of value. Due to the recent popularity and trade of self-driving cars and driver assistance systems, traffic sign recognition (see Figure 1) gained considerable attention among machine vision tasks~\cite{sanyal2020traffic}. Conversely, deep learning techniques are now the dominant solution for machine learning and especially machine learning issues. Naturally, machine vision became the epitome of deep learning.
Convolutional Neural Network (CNN) is one of the most well-known architectures with a remarkable ability to maintain local dependencies in deep learning realm. Their filters perceive a local region of the input image. In other words, CNNs are very efficient in learning local features. These Networks performed well in various machine vision applications and achieved excellent results.

After the success of CNNs, transformers were introduced. Before machine vision transformers became the state-of-the-art(SOTA) solution in Natural Language Processing (NLP) studies. After that, researchers began to modify transformers and use them in machine vision tasks. This approach led to the introduction of vision transformers (ViT)~\cite{Vit}. Unlike CNNs, these transformers do not detect correlations in nearby pixels very well. But they are very efficient in learning global dependencies by maintaining the relationship between every pixel through the image. First, basic Vits achieved encouraging results in machine vision tasks, but they required a high amount of data for training and still needed help to reach SOTA CNN models. So researchers tried to modify the primary vision transformer’s architecture and introduced more complicated models~\cite{Dilated} which performed much better than basic ones and even outperformed some of the best convolutional networks.

\begin{figure}[!t]
 \centering
  \includegraphics[width=.9\linewidth]{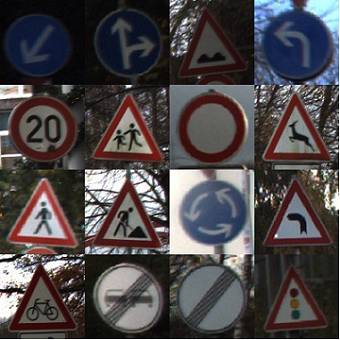}
  \caption{
The provided images display various traffic signs sourced from the GTSRB dataset~\cite{GTSRB}.}\label{fig.GTSRB}
\vspace{-4mm}
\end{figure}

As local and global features play a vital role in classification tasks, researchers have begun using convolutions and transformers in their newly designed hybrid models. These models outperformed pure transformer-based models, and some of them outperformed state-of-the-art convolutional models as well.
So, we introduced a new hybrid model to answer shortcuts of convolutional and pure transformer-based networks and use both advantages. Our proposed model consists of convolutional and transformer-based blocks. Convolutional blocks are used for perceiving and learning the correlation between near pixels. Transformer-based blocks are used for sensing global relationships. We also added a locality module in one of the transformer-based blocks to add more local perception to the model.

So far, our contributions in this work are: 1- We proposed a new hybrid model and added a locality module. 2- We analyzed our model’s performance on German Traffic Sign Recognition Benchmark (GTSRB)~\cite{GTSRB} and Persian Traffic Sign Dataset (PTSD)~\cite{PTSD}. We compared it with other SOTA convolutional and transformer-based models. 3- We used AugMix~\cite{hendrycks2019augmix} data augmentation to add more robustness to the model, which raised accuracy.
This paper has been arranged in the following order: Section 2 investigates previous studies on TSR and classification tasks. Section 3 explains applied modification both in terms of architecture and computation. In section 4, we will demonstrate and analyze our results. The results contain all previous models and the newly modified model’s performance in both datasets. Lastly, in the final part, we summarize the findings of our work and make suggestions for further research.

 \begin{figure*}[!t]
 \centering
  \includegraphics[width=.8\textwidth]{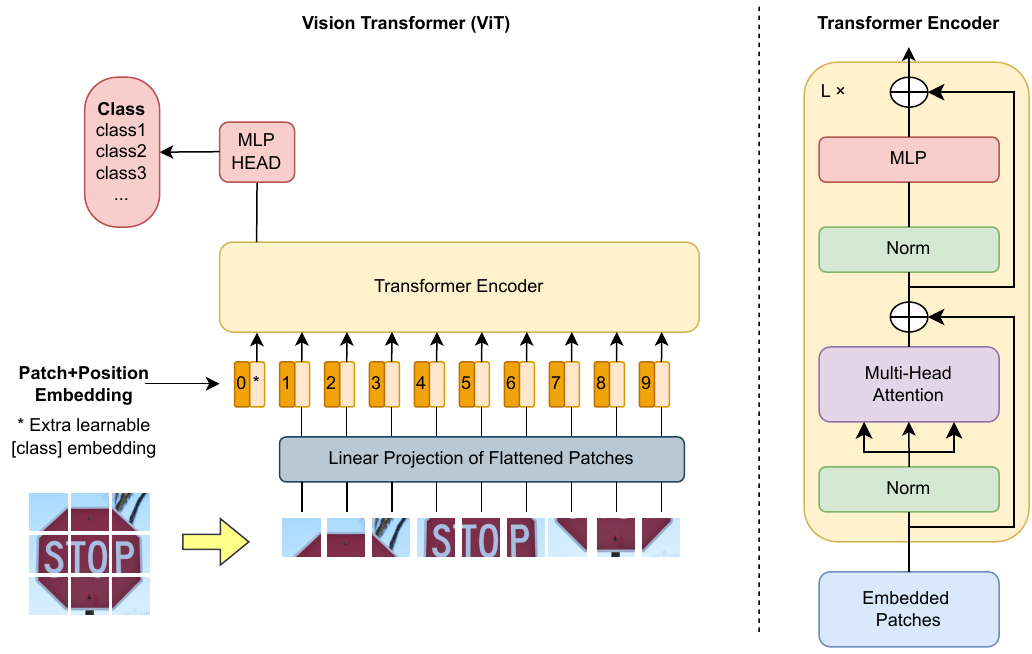}
  \caption{Network architecture of the vanilla ViT.}\label{fig.ViT}
  \vspace{-4mm}
\end{figure*}

\section{Related Works}

Various methods have been explored for developing traffic sign recognition systems. However, since 2013, using convolutional neural networks has become standard in object recognition tasks due to their capability of learning a hierarchy of features by producing low-level features from high-level features~\cite{arcos2018evaluation}. In~\cite{Embeded}, the author has proposed a committee of convolutional layers for feature extraction. This committee enhances the model’s ability to identify patterns within the image. This model achieved an accuracy of 98.41\% with only 2.61M parameters. In~\cite{haloi2015traffic}, another convolutional model was introduced by designing an inception module that can attend to the entire image. This model is highly accurate and has a relatively small number of parameters. In recent work\cite{nezhad2021transfer}, ensemble learning was used to recognize traffic signs. Ensemble learning involves using multiple models simultaneously and in combination for decision-making. In this study, pre-trained DensNet~\cite{huang2017densely}, VGG~\cite{VGG16} and ResNet~\cite{Resnet} models were utilized. Additionally, the number of data samples was significantly increased, with the amount of images in the GTSRB dataset rising from 51,000 to 96,000. The proposed model was found to be more accurate than other models in the comparison made in this study. However, it did not achieve 99\% accuracy.

After the success of Transformers in the NLP area, researchers tried using them in vision tasks. The first severe vision transformer introduced was VIT~\cite{Vit}. After VIT, some other models were introduced, mostly VIT’s improvements. The work~\cite{LNL} presents a comparison between convolutional models and vision transformers in the field of TSR. The study evaluated seven convolutional models, such as DensNet, VGG, ResNet and MobileNet and five vision transformers like VIT and PVT~\cite{pyramid}. The results showed that advanced convolutional models outperformed primary vision transformers. Particularly, the test phase accuracy on the GTSRB dataset for the VGG model was the greatest. In  incorporation of convolution inductive biases into the early blocks of transformer led to outperforming SOTA models. In~\cite{medvit}, newer vision transformer models, such as CCT~\cite{hassani2021escaping}, which used convolution to build image segments, were compared with DensNet for traffic sign recognition. This study showed that not all vision transformer models are suitable for this task. The comparison results demonstrated that pre-trained DensNet was superior. However, the CCT model was able to outperform the untrained DensNet model.

\section{METHODOLOGY}

In this section, we will introduce our proposed model. Before explaining its architecture, we will briefly overview the vision transformer’s architecture and equations to better understand the proposed model. After that, we will describe our model and its mechanism.

\subsection{Overview of Vision Transformer}

The primary transformer architecture contains of a decoder and an encoder, which have similar structures. Since we are focused on the classification of traffic signs, only the encoder portion of the transformer is required. The encoder typically comprises two parts.As illustrated in Figure 2 The first part includes a self-attention mechanism that evaluates the relationship between each token and all the other tokens. The second part is a simple fully-connected neural network with two layers that extracts more and richer features from each token. From a mathematical perspective, since transformers operate on one-dimensional data, a two-dimensional input image with dimensions of C × H × W must be converted into a sequence of tokens. less mathematical calculations are required, so we first divide the input image into P × P patches, resulting in a set of images with dimensions P × P × C. The number of this set equals:

\begin{equation}\label{eq.1}
 N = (H \times W) / P^2
\end{equation}

Also, the size of each token is:

\begin{equation}\label{eq.2}
 d = C \times P^2
\end{equation}

However, this number can be very high. If we use this sequence directly in the transformer, we would need to multiply a matrix with dimensions N×d by a matrix with dimensions d × N, which increases the computational complexity and memory consumption of the model and thus decrease the speed. To address this issue, linear mapping reduces the tokens' dimensionality to the model's the desired number. For instance, if the desired number is D, the input matrix will be multiplied by a matrix with dimensions N × D. A convolutional layer or a simple MLP achieves linear mapping in practice.

The input tokens are fed into the self-attention mechanism. First, Key (K), Query (Q), and Value (V) matrices are created through mapping from the input using fully connected networks (FCs). The dimensions of these matrices are identical to the input dimensions (N × D). Next, the self-attention mechanism applies relation 3 to these matrices.

\begin{equation}
\mathrm{Z}=\operatorname{softmax}\left(\frac{\mathbf{Q}K^T}{\sqrt{D}}\right) \mathbf{V}
\end{equation}

The output spatial dimensions of the self-attention mechanism remain the same as the input dimensions. The resulting output is passed to the feed-forward section after the addition of the normalized sum. The FFN section mainly comprises of two fully connected (FC) layers that extract richer and more complex features. This is obtained by growing the amount of neurons in the first layer and multiplying the dimension of features. On the other hand, the second layer has the same number of neurons as the input, ensuring that the dimensions of the input and output of the feed-forward section remain the same. This approach is similar to the residual block in ResNet~\cite{Resnet}.

\subsection{Next Generation of Local Vision Transformer}
The objective of our model is to extract both local features of an image. To achieve this, we suggest a novel hybrid network that combines convolutional and transformer layers. The convolutional are particularly effective for capturing high-frequency features such as edges and corners, while transformers are better suited for capturing low-frequency features such as global patterns. The Next-ViT~\cite{li2022next} model inspires our model and includes NCB and NTB blocks. We will describe the architecture and equations of these blocks and then discuss the strategy for arranging them in the model called NHS.

 \begin{figure*}[!t]
 \centering
  \includegraphics[width=.7\textwidth]{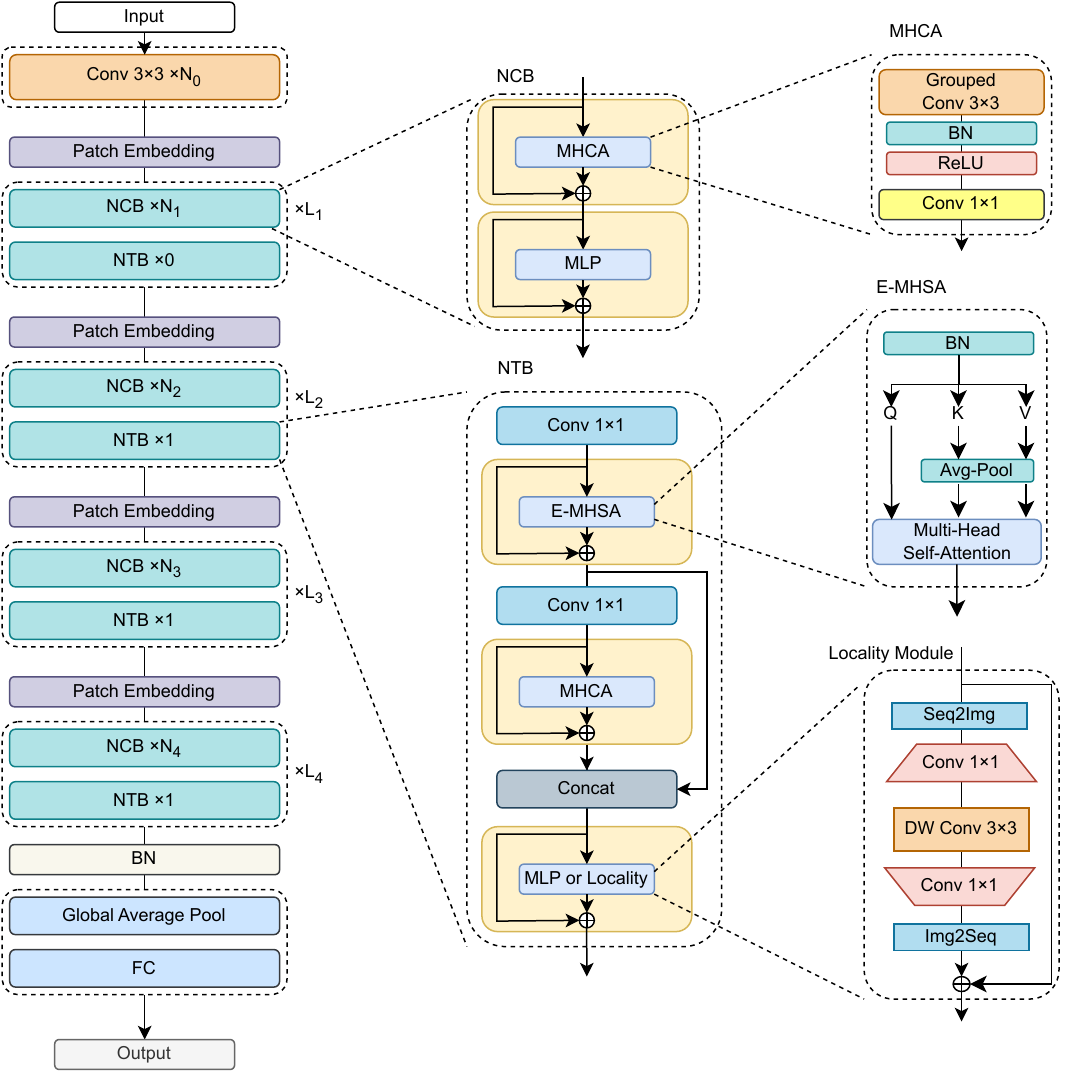}
  \caption{The architecture of proposed Next Generation of Local Vision Transformer.}\label{fig.architecure}
  \vspace{-4mm}
\end{figure*}

\subsection{Next Convolution Block}

Our goal with the NCB block is to leverage convolutional layers while maintaining the high performance of transformers. A token mixer is utilized to achieve this, emphasizing the importance of Multi-Head Convolutional Attention (MHCA). We shall go into great depth about MHCA in the section that follows. The NCB block consists of one MHCA and one MLP, which together can be described by the following equations:
$$
\begin{aligned}
& \hat z^{l} &&= MHCA(z^{l-1}) + z^{l-1} \\
& z^{l} &&= MLP(\hat z^{l}) + \hat z^{l}
\end{aligned}
$$

Here $z^{l-1}$ represents the input from the $l-1$ component, while $\hat z^{l}$ and $z^l$ represent the output of the MHCA and the $lth$ NCB.

\subsubsection{\textbf{MHCA}}

To improve the speed of the attention-based token mixer, we introduce an attention mechanism with the convolutional operation named Convolutional Attention (CA). Furthermore, inspired by basic transformers, we utilize a multi-head architecture to attend to features from various representation subspaces, which allows for efficient local data learning.  The general functioning of MHCA can be summarized as bellow:

\begin{equation}
\operatorname{MHCA}(x) =\operatorname{Concat}\left(\mathrm{CA}_1\left(x_1\right), \mathrm{CA}_2\left(x_2\right), \ldots, \mathrm{CA}_h\left(x_h\right)\right) W^P
\end{equation}

In this equation, the Multi-Head Channel Attention (MHCA) mechanism is employed to extract information from the divided feature z, which is split into h heads. To further
enhance the interaction of information from multiple heads, a projection layer (WP) is utilized. Single-head convolutional attention, or CA, is referred to in this context and is defined as:

\begin{equation}
\mathrm{CA}(X) = O \left(\mathrm{W} \cdot \mathrm{T}_{\{i,j\}}\right), \text { where } \mathrm{T}_{\{i,j\}} \in X
\end{equation}

In this equation, $T_i$ and $T_j$ represent neighboring tokens within the input feature $z$. $O$ denotes an inner product operation with $W$ as a learnable parameter and input patches. Through iterative optimization of parameter $W$, $O$ can effectively learn the connectivity between various tokens within the local receptive field.


\subsection{Next Transformer Block}

Accurate recognition requires the capture of both local and global information. While NCB has demonstrated proficiency in learning local features, we have introduced an additional block to capture global features. Transformer blocks are known for their effectiveness in capturing low-frequency signals, which provide global information, but they are not as proficient in capturing high-frequency information. To address this issue, we have produced the Next Transformer component, which operates in a multi-frequency manner to capture both high-frequency and low-frequency signals. The first step in our approach involves the use of a novel mechanism called Efficient Multi-Head Self Attention (E-MHSA), that could be formulated as follows:

\begin{equation}
  \operatorname{ESA}(x) =\operatorname{Concat}\left(\operatorname{SA}_{1}\left(x_{1}\right), \mathrm{SA}_{2}\left(x_{2}\right), \ldots, \mathrm{SA}_{h}\left(x_{h}\right)\right) W^{O}
\end{equation}

Where z represents the input feature like before, and SA means spatial reduction self-attention operator, which performs as:

\begin{equation}
  \operatorname{SA}(X) =\operatorname{Attention}\left(X \cdot W^{Q}, \mathrm{P}_{s}\left(X \cdot W^{K}\right), \mathrm{P}_{s}\left(X \cdot W^{V}\right)\right)
\end{equation}

Here, the term Attention refers to routine calculation as Eq(3). $W^Q$, $W^K$, and $W^V$ are linear layers, and $P_s$ is an avg-pool operator with a stride of $S$ to reduce computational cost. It is clear that the time required for the E-MHSA is directly proportional to the number of channels it has. NTB utilizes a point-wise convolution to expedite the process by reducing the channel dimension. A shrinking ratio, denoted as r, is introduced to achieve this. In addition, NTB incorporates an MHCA module that works the E-MHSA to extract multi-frequency features. Following MHCA, both MHCA and E-MHSA outputs are concatenated to blend high- and low-frequency data information. An MLP is often used to extract key characteristics near the ending of a neural network. However, following the approach presented in~\cite{li2021localvit}, we utilize a Local Feed-Forward (LFF) module exclusively in the first NTB block, while MLP is still utilized in the rest of the NTB blocks. The locality module is an inverted residual block~\cite{sandler2018mobilenetv2} that employs convolutions, thereby enhancing the local vision of the network as compared to a simple MLP. The locality module in the first NTB block is used because the initial blocks focus on capturing local features, such as high-frequency features. The complete implementation of NTB (excluding the first block for the last equation) can be calculated as bellow:

\begin{equation}
\begin{aligned}
\bar{z}^{l} &=\operatorname{Proj}\left(z^{l-1}\right) \\
\ddot{z}^{l} &=\mathrm{E-MHSA}\left(\bar{z}^{l}\right)+\bar{z}^{l} \\
\dot{z}^{l} &=\operatorname{Proj}\left(\ddot{z}^{l}\right) \\
\tilde{z}^{l} &=\operatorname{MHCA}\left(\dot{z}^{l}\right)+\dot{z}^{l} \\
\hat{z}^{l} &=\operatorname{Concat}\left(\ddot{z}^{l}, \tilde{z}^{l}\right) \\
z^{l} &=\operatorname{LFF}\left(\hat{z}^{l}\right)+\hat{z}^{l}
\end{aligned}
\end{equation}

\begin{table}[t]
    \caption{Achievement of SOTA models for Traffic Sign Recognition on the GTSRB.}
    \begin{adjustbox}{width=.95\linewidth,center}

    \begin{tabular}{l|ccc}
    \toprule
         Method & Accuracy (\%) & Parameters (M) & GFLOPS (G) \\

        \midrule EfficientNetV2 & 99.33 & 20.23 & 8.48 \\
        \hline InceptionResnetV2 & 99.11 & 54.37 & 6.54 \\
        \hline ResNet34 & 98.94 & 21.31 & 3.68  \\
        \hline DenseNet121 & 98.9 & 7.00 & 2.86  \\
        \hline Next-Vit & 99.35 & 30.78 & 5.8  \\
        \hline PIT & 99.18 & 22.9 & 2.87 \\
        \hline VITAE-S & 99.15 & 23.66 & 6.2 \\
        \hline PVT-V2 & 98.89 & 24.87 & 4.04 \\
        \hline VITAE-6m & 98.64 & 6.37 & 2.17 \\
        \hline VIT-b16 & 98.57 & 86.57 & 17.56\\
        \hline R50+VIT-b-16 & 98.84 & 97.9 & 22.36\\
        \hline Next-LVT (ours) & \textbf{99.66} & \textbf{33.14} & \textbf{7.66}\\
        \bottomrule
        \end{tabular}
    \end{adjustbox}
    \vspace{-5mm}
\end{table}

\subsubsection{Next Hybrid Strategy}

Traditionally, convolutions are frequently used in shallow phases of conventional hybrid models, and transformer blocks are stacked in the final stage or stages. However, this approach can fail to capture global information in the early stages, resulting in suboptimal performance. We propose a new method called Next Hybrid Strategy (NHS) to address this issue. The NHS strategy follows an (NCB×N+NTB×1) pattern, where each stage has one $NTB$ and $N NCB$. To allow the model to learn global features in shallow layers, the transformer block is explicitly included at the end of each step. Finally, we repeat each stage by the factor of $L$. So the final pattern of the model will be $(NCB×N+NTB×1)×L$, as shown in Figure \ref{fig.architecure}.

\section{Experiments}

In this part, we initiate the discussion by examining the datasets utilized to train our enhanced Transformer model, and provide a summary of the experimental configurations. Following that, we conduct a comprehensive analysis of the experimental outcomes, wherein We evaluate the effectiveness of our Transformer model with recent breakthroughs in the field of traffic sign recognition.

\subsection{Datasets and Implementation details}

After discussing the model’s architecture, we will now shift focus to the experiments. In total, we conducted 23 experiments on two datasets: GTSRB and PTSD. We trained and tested various models on these datasets. We ran our experiments using Google Colab, which provided us with a Tesla T4 GPU and 15 gigabytes of RAM. We implemented our experiments using the PyTorch framework. Now our training and testing setup is as follows. With a linear learning scheduler that split the learning by $10$ after three training epochs, we set the starting learning rate at $0.007$. For the GTSRB dataset, we trained and tested for 20 epochs, and For the PTSD dataset, we trained and tested for 15 epochs. Our test batch size was set at 256 and our training batch size to 64. We used the Cross-Entropy loss function and the SGD optimizer to determine the loss during training. As for our dataset’s setup, the respective dataset sources provided the train and test sets. For the GTSRB dataset, 39,209 training photos and 12,630 test images totaling 43 distinct classes of traffic signs were used. For the PTSD dataset, we had a total of 14,405 training images and 2,421 test images, also containing 43 different types of traffic signs. It is worth noting that the amount of images in the train and test sets varied between the two datasets. We also employed AugMix~\cite{hendrycks2019augmix} as our primary data augmentation technique to enhance the robustness of our model.

\begin{table}[t]
    \caption{Performance evaluation of suggested Transformer and SOTA models for PTSD.}
    \begin{adjustbox}{width=.9\linewidth,center}

        \begin{tabular}{l|ccc}
        \toprule
         Method & Accuracy (\%) & Parameters (M) & GFLOPS (G) \\

        \midrule EfficientNetV2 & 99.67 & 20.23 & 8.48 \\
        \hline InceptionResnetV2 & 99.46 & 54.37 & 6.54 \\
        \hline DenseNet121 & 99.67 & 7.00 & 2.86  \\
        \hline Next-Vit & 99.7 & 30.78 & 5.8  \\
        \hline PIT & 99.67 & 22.9 & 2.87 \\
        \hline VITAE-S & 99.3 & 23.66 & 6.2 \\
        \hline PVT-V2 & 99.55 & 24.87 & 4.04 \\
        \hline VITAE-6m & 99.5 & 6.37 & 2.17 \\
        \hline Next-LVT (ours) & \textbf{99.8} & \textbf{33.14} & \textbf{7.66}\\
        \bottomrule
        \end{tabular}
    \end{adjustbox}
    \vspace{-5mm}
\end{table}

\subsection{RESULTS}

In Table.1, we present the results of our experiments on the GTSRB dataset, including the accuracy, whole amount of parameters, and total amount of GFLOPs for each of the 14 models. Table.2 presents the same metrics for our experiments on the PTSD dataset. These metrics provide valuable information about our model's computational complexity and efficiency and allow easy comparison between different models.
Table 1 shows that our model, called Next-generation Local Vision Transformer (Next-LVT), achieved an accuracy of 99.66\%, surpassing other convolutional, transformer, and hybrid models. While our model has a slightly higher number of total parameters and GFLOPs, it remains suitable for real-time applications. Similarly, Table 2 demonstrates that our model achieved the highest accuracy of 99.8\%. The differences in the PTSD dataset were smaller than those in the GTSRB dataset, indicating that PTSD is a less complex and diverse dataset.

\section{Conclusion}

This paper addresses the importance of safety systems in self-driving cars and driver assistance systems, specifically on traffic sign recognition. We suggest a new novel model that combines the strong points of both CNNs and transformers. Our model consists of convolutional and transformer-based blocks, leveraging the advantages of each approach. A locality module is also introduced to excel the model's local perception. The suggested model is evaluated on the GTSRB and PTSD and compared against SOTA convolutional and transformer architectures. The results demonstrate that our hybrid network outperforms pure transformer-based models and surpasses some of the best convolutional networks. The proposed model achieves improved accuracy in traffic sign recognition tasks by incorporating local and global features. Furthermore, using data augmentation techniques, such as AugMix, enhances the model's robustness. Future research directions may involve further improvements to the hybrid model and exploring its application in other computer vision tasks.

\bibliographystyle{IEEEtran}

{\small
\bibliography{article}}

\end{document}